# Natural Language Inference for Arabic Using Extended Tree Edit Distance with Subtrees

**Maytham Alabbas**                                       MAYTHAM.ALABBAS@GMAIL.COM
*Department of Computer Science, University of Basrah,*
*Basrah, Iraq*

**Allan Ramsay**                                      ALLAN.RAMSAY@MANCHESTER.AC.UK
*School of Computer Science, University of Manchester,*
*Manchester, M13 9PL, UK*

## Abstract

Many natural language processing (NLP) applications require the computation of similarities between pairs of syntactic or semantic trees. Many researchers have used tree edit distance for this task, but this technique suffers from the drawback that it deals with single node operations only. We have extended the standard tree edit distance algorithm to deal with subtree transformation operations as well as single nodes. The extended algorithm with subtree operations, TED+ST, is more effective and flexible than the standard algorithm, especially for applications that pay attention to relations among nodes (e.g. in linguistic trees, deleting a modifier subtree should be cheaper than the sum of deleting its components individually). We describe the use of TED+ST for checking entailment between two Arabic text snippets. The preliminary results of using TED+ST were encouraging when compared with two string-based approaches and with the standard algorithm.

## 1. Introduction

Tree edit distance has been widely used as a component of natural language processing (NLP) systems that attempt to determine whether one text snippet supports an inference to another (roughly speaking, whether the first *entails* the second), with the distance between pairs of dependency trees being taken as a measure of the likelihood that one entails the other. We extend the standard algorithm for calculating the distance between two trees by allowing operations to apply to subtrees, rather than just to single nodes. This extension improves the performance of our technique for Arabic by around 5% in F-score and around 4% in accuracy compared with a number of well-known techniques. The relative performance of the standard techniques on our Arabic testset replicates the results reported for these techniques for English testsets. We have also applied our extended version of tree edit distance, TED+ST, to the English RTE-2 testset, where it again outperforms the standard algorithm.

Tree edit distance is a generalisation of the standard string edit distance metric, which measures the similarity between two strings. It has been used to underpin several NLP applications such as information extraction (IE), information retrieval (IR) and natural language inference (NLI). The edit distance between two trees is defined as the minimum cost sequence of edit operations to transform one tree to another. There have been numerous approaches to calculating edit distance between trees, as reported by Selkow (1977), Tai (1979), Zhang and Shasha (1989), Klein (1998), Demaine, Mozes, Rossman, and Weimann (2009) and Pawlik and Augsten (2011). We have chosen to work with Zhang-Shasha's algorithm (Zhang & Shasha, 1989) because the intermediate structures produced by this algorithm allow us to detect and respond to operations on subtrees. When we refer to the standard tree edit distance algorithm throughout the rest of this article, we mean Zhang-Shasha's algorithm, for which we will use the short form ZS-TED.

Our ultimate goal is to develop an NLI system for Arabic (Alabbas, 2011).[1] NLI is the problem of determining whether a natural language hypothesis $h$ can reasonably be inferred from a natural language premise $p$. The challenges of NLI are quite different from those encountered in formal deduction: the emphasis is on informal reasoning, lexical semantic knowledge, and variability of linguistic expression, rather than on long chains of formal reasoning (MacCartney, 2009). A more recent, and better-known, formulation of the NLI task is the *recognising textual entailment challenge* (RTE), described by Dagan and Glickman (2004) as a task of determining, for two text snippets premise $p$ and hypothesis $h$, whether "...typically, a human reading $p$ would infer that $h$ is most likely true." According to these authors, entailment holds if the truth of $h$, as interpreted by a typical language user, can be inferred from the meaning of $p$. A popular method that has been used in recent years for such tasks is the use of tree edit distance, which compares sentence pairs by finding a minimal cost sequence of editing operations to transform a tree representation of one sentence into a tree for the other (Kouylekov, 2006; Heilman & Smith, 2010). Approximate tree matching of this kind allows users to match parts of two trees, rather than demanding a complete match of every element of each tree. However, one of the main drawbacks of tree edit distance is that transformation operations are applied solely on single nodes (Kouylekov, 2006). Kouylekov and Magnini (2005) used the standard tree edit distance, which uses transformation operations (insert, delete and exchange) solely on single nodes, to check the entailment between two dependency trees. On the other hand, Heilman and Smith (2010) extended the available operations in standard tree edit distance to `INSERT-CHILD`, `INSERT-PARENT`, `DELETE-LEAF`, `DELETE-&-MERGE`, `RELABEL-NODE` and `RELABEL-EDGE`. These authors also identify three new operations, `MOVE-SUBTREE`, which means move a node $X$ in a tree $T$ to be the last child of a node $Y$ in $T$ (s.t. $Y$ is not a descendant of $X$), `NEW-ROOT` and `MOVE-SIBLING`, to enable succinct edit sequences for complex transformation. This extended set of edit operations allows certain combinations of the basic operations to be treated as single steps, and hence provides shorter (and therefore cheaper) derivations. The fine-grained distinctions between, for instance, different kinds of insertions also make it possible to assign different weights to different variations on the same operation. Nonetheless, these operations continue to operate on individual nodes rather than on subtrees (despite its name, even `MOVE-SUBTREE` appears to be defined as an operation on nodes rather than on subtrees). We have solved this problem by extending the basic version of the algorithm so that the costs of operations that insert/delete/exchange subtrees are derived by some appropriate function of the costs of the operations on their parts. This makes TED+ST more effective and flexible than the standard algorithm, especially for applications that pay attention to relations among nodes (e.g deleting a modifier subtree, in linguistic trees, should be cheaper than the sum of deleting its components individually).

---

1. In particular, for Modern standard Arabic (MSA). When we refer to Arabic throughout this article, we mean MSA.

The rest of the paper is organised as follows: Zhang-Shasha's algorithm, ZS-TED, is explained in Section 2. Section 3 presents TED+ST. Section 4 describes dependency trees matching. Dataset preparation is explained in Section 5. The experimental results are discussed in Section 6. Conclusions are given in Section 7.

## 2. Zhang-Shasha's TED Algorithm

Our approach extends ZS-TED, which uses dynamic programming to provide an $O(n^4)$ algorithm for finding the optimal sequence of node-based edit operations for transforming one tree into another. This section contains a brief recapitulation of this algorithm–a more detailed description is given by Bille (2005).

Ordered trees are trees in which the left-to-right order among siblings is significant. Approximate tree matching allows us to match a tree with just some parts of another tree. There are three operations, namely deleting, inserting and exchanging a node, which can transform one ordered tree to another. There is a nonnegative real cost associated with each edit operation. These costs are changed to match the requirements of specific applications. Deleting a node $x$ means attaching its children to the parent of $x$. Insertion is the inverse of deletion. This means an inserted node becomes a parent of a consecutive sub-sequence in the left to right order of its parent. Exchanging a node alters its label. All these editing operations are illustrated in Figure 1 (Bille, 2005).

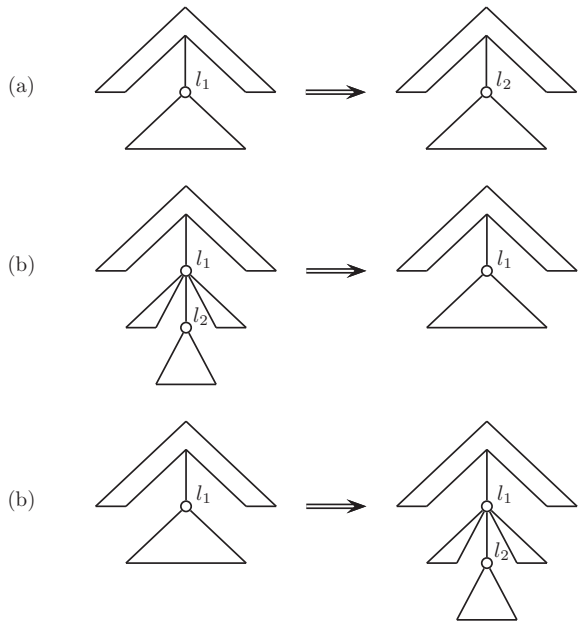

Figure 1: (a) Relabeling the node label ($l_1 \rightarrow l_2$). (b) Deleting the node labeled ($l_2 \rightarrow \wedge$). (c) Inserting a node labeled $l_2$ as the child of the node labeled $l_1$ ($\wedge \rightarrow l_2$).

Each operation is associated with a cost and is allowed on single nodes only. Selecting a good set of costs for these operations is hard when dealing with complex problems. This is because alterations in these costs or choosing a different combination of them can lead to drastic changes in tree edit distance performance (Mehdad & Magnini, 2009).

In the ZS-TED algorithm, tree nodes are compared using a postorder traversal, which visits the nodes of a tree starting with the leftmost leaf descendant of the root and proceeding to the leftmost descendant of the right sibling of that leaf, the right siblings, and then the parent of the leaf and so on up the tree to the root. The last node visited will always be the root. An example of the postorder traversal and the leftmost leaf descendant of a tree is shown in Figure 2. In this figure, there are two trees, $T_1$ with $m{=}7$ nodes and $T_2$ with $n{=}7$ nodes. The subscript for each node is considered the order of this node in the postorder of the tree. So, the postorder of $T_1$ is $e,f,b,g,c,d,a$ and the postorder for $T_2$ is $g,c,y,z,x,d,a$. The leftmost leaf descendant of the subtrees of $T_1$ headed by the nodes $e,f,b,g,c,d,a$ are 1,2,1,4,4,6,1 respectively, and similarly the leftmost leaf descendants of $g,c,y,z,x,d,a$ in $T_2$ are 1,1,3,4,3,3,1.

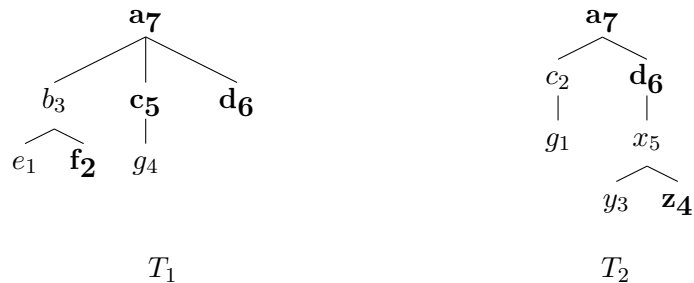

Figure 2: Two trees $T_1$ and $T_2$ with their postorder traversal.

For all the descendants of each node, the least cost mapping has to be calculated before the node is encountered, in order that the least cost mapping can be selected right away. To achieve this, the algorithm pursues the keyroots of the tree, which are defined as a set that contains the root of the tree plus all nodes having a left sibling. Concentrating on the keyroots is critical to the dynamic nature of the algorithm, since it is the subtrees rooted at keyroots that allow the problem to be split into independent subproblems of the same general kind. The keyroots of a tree are decided in advance, permitting the algorithm to distinguish between *tree distance* (the distance between two nodes when considered in the context of their left siblings in the trees $T_1$ and $T_2$) and *forest distance* (the distance between two nodes considered separately from their siblings and ancestors but not from their descendants) (Kouylekov, 2006). For illustration, the keyroots in each tree in Figure 2 are marked in bold.

For each node, the computation to find the least cost mapping (the tree distance) between a node in the first tree and one in the second depends solely on mapping the nodes and their children. To find the least cost mapping of a node, then, one needs to recognise the least cost mapping from all the keyroots among its children, plus the cost of its leftmost child. Because the nodes are numbered according to the postorder traversal, the algorithm proceeds in the following steps (Kouylekov, 2006): (i) the mappings from all leaf keyroots are determined; (ii) the mappings for all keyroots at the next higher level are decided recursively; and (iii) the root mapping is found. Algorithm 1 shows the pseudocode of ZS-TED algorithm (Zhang & Shasha, 1989). The matrices $D$ and $FD$ are used for recording the results of individual subproblems: $D$ is used to store the tree distance between trees rooted at pairs of nodes in the two trees, and $FD$ is used to store the 'forest distance' between *sequences of nodes*. $FD$ is used as a temporary store while the tree edit distance between pairs of keyroots are being calculated. We have extended the standard algorithm, which computes the *cost* of the cheapest edit sequence, so that it also records the edit operations themselves. This involves adding two new matrices, $DPATH$ and $FDPATH$, to hold the appropriate sequences of edit operations–$DPATH$ to hold the edit sequences for trees rooted at pairs of nodes and $FDPATH$ to hold the edit sequences for forests. $D$ and $DPATH$ are permanent arrays, whereas $FD$ and $FDPATH$ are reinitialised for each pair of keyroots.

The algorithm iterates over keyroots, and is split into two main stages for each pair of keyroots: the initialisation phase (lines 3–12) deals with the first row and column, where we assume that every cell in the first row is reached by appending the insert operation "i" to the cell to its left and every cell in the first column is reached by appending the delete operation "d" to the cell above it, with appropriate costs. This is exactly parallel to the initialisation of the standard dynamic time warping algorithm for calculating string edit distance, as though we were treating the task of matching the subsets of the subtrees rooted at $T_1[x]$ and $T_2[y]$ as a string matching problem between the nodes in these two trees as sequences enumerated in post-order.

The second stage (lines 13–37) traces the cost and edit sequence for transforming each sub-sequence of the sequence of nodes dominated $T_1[x]$ to each sub-sequence of the sequence of nodes dominated $T_2[x]$, by considering whether the nodes in these were reached from the cell to the left by an insert, or from the cell above by a delete, or by the cell diagonally above and left by either a match "m" or an exchange "x." There are two cases to be considered here:

i If the two sequences under consideration are both trees (tested at line 15), then we know that we have considered every possible way of exchanging one into the other, and hence we can record the cost in both $FD$ and $D$, and the edit sequence in both $FDPATH$ and $DPATH$. In this case, we calculate the cost of moving along the diagonal by inspection of the two nodes. See Figure 3 for an illustration of this notion.

ii If one or both of the sequences is a forest we retrieve the cost of moving along the diagonal from $DPATH$, and we just store the cost in $FD$ and the edit sequence in $FDPATH$.

In both cases, we gather the set of $\{cost, path\}$ pairs that result from considering insert/delete/exchange operations on the preceding sub-sequences, and choose the best such pair to store in the various arrays. This is again very similar to the corresponding element of the string edit algorithm, with the added complication that calculating the tree edit costs and sequences for a pair of keyroots involves calculating the costs and edit sequences for all pairs of sub-sequences of the nodes below those roots. The results for pairs of keyroots are stored permanently, and are utilised during the calculations for sub-sequences at the next stage.

Bille (2005) provides detailed worked examples of the calculation of the costs of transforming one tree into another. Figure 4 shows how $FDPATH$ grows as the algorithm iterates through the keyroots for the trees $T_1$ and $T_2$ in Figure 2. In this figure, the cells representing

**Algorithm 1** pseudocode for Zhang-Shasha's TED algorithm with edit sequences

| | |
|---|---|
| $T[i,j]$ | $i_{th}$ to $j_{th}$ nodes in the post-order enumeration of tree $T$ ($T[i,i]$ is written $T[i]$) |
| $l(i)$ | the leftmost leaf descendant of the subtree rooted at i |
| $K(T)$ | the keyroots of tree $T$, $K(T) = \{k \in T : \neg \exists k_1 > k$ with $l(k_1) = l(k)\}$ |
| $D[i,j]$ | the tree distance between two nodes $T_1[i]$ and $T_2[j]$ |
| $FD[T_1[i,i_1], T_2[j,j_1]]$ | the forest distance from nodes $i$ to $i_1$ in $T_1$ to nodes $j$ to $j_1$ in $T_2$ |
| $DPATH[i,j]$ | edit sequence for trees rooted at two nodes $T_1[i]$ and $T_2[j]$ |
| $FDATH[T_1[i,i_1], T_2[j,j_1]]$ | edit sequence for forests covered by nodes $i$ to $i_1$ in $T_1$ to nodes $j$ to $j_1$ in $T_2$ |
| $\gamma(T_1[i] \longrightarrow \wedge)$ | cost of deleting the $i_{th}$ node from $T_1$ |
| $\gamma(\wedge \longrightarrow T_2[j])$ | cost of inserting the $j_{th}$ node of $T_2$ into $T_1$ |
| $\gamma(T_1[i] \longrightarrow T_2[j])$ | cost of exchanging the $i_{th}$ node of $T_1$ with the $j_{th}$ node of $T_2$ |
| $m, n$ | the number of nodes in $T_1$ and $T_2$ respectively |
| $best$ | choose the best cost and path from a set of options |

1: **for** $x \leftarrow 1$ **to** $|K_1(T_1)|$ **do**
2:     **for** $y \leftarrow 1$ **to** $|K_2(T_2)|$ **do**
3:         $FD[\emptyset, \emptyset] \leftarrow 0$
4:         $FDPATH[\emptyset, \emptyset] \leftarrow$ " "
5:         **for** $i \leftarrow l_1(x)$ **to** $x$ **do**
6:             $FD[T_1[l_1(x),i], \emptyset] \leftarrow FD[T_1[l_1(x),i\text{-}1], \emptyset] + \gamma(T_1[i] \longrightarrow \wedge)$
7:             $FDPATH[T_1[l_1(x),i], \emptyset] \leftarrow FDPATH[T_1[l_1(x),i\text{-}1], \emptyset] +$ "d"
8:         **end for**
9:         **for** $j \leftarrow l_2(y)$ **to** $y$ **do**
10:             $FD[\emptyset, T_2[l_2(y),j]] \leftarrow FD[\emptyset, T_2[l_2(y),j\text{-}1]] + \gamma(\wedge \longrightarrow T_2[j])$
11:             $FDPATH[\emptyset, T_2[l_2(y),j]] \leftarrow FDPATH[\emptyset, T_2[l_2(y),j\text{-}1]] +$ "i"
12:         **end for**
13:         **for** $i \leftarrow l_1(x)$ **to** $x$ **do**
14:             **for** $j \leftarrow l_2(y)$ **to** $y$ **do**
15:                 **if** $(l_1(i) == l_1(x)$ **and** $l_2(j) == l_2(y))$ **then**
16:                     $cost, path \leftarrow best(\{FD[T_1[l_1(x),i\text{-}1], T_2[l_2(y),j]] + \gamma(T_1[i] \longrightarrow \wedge),$
17:                             $FDPATH[T_1[l_1(x),i\text{-}1], T_2[l_2(y),j]] +$ "d"$\},$
18:                             $\{FD[T_1[l_1(x),i], T_2[l_2(y),j\text{-}1]] + \gamma(\wedge \longrightarrow T_2[j]),$
19:                             $FDPATH[T_1[l_1(x),i], T_2[l_2(y),j\text{-}1]] +$ "i"$\},$
20:                             $\{FD[T_1[l_1(x),i\text{-}1], T_2[l_2(y),j\text{-}1]] + \gamma(T_1[i] \longrightarrow T_2[j])),$
21:                             $FDPATH[T_1[l_1(x),i\text{-}1], T_2[l_2(y),j\text{-}1]] +$ "m"/"x"$\})$
22:                 $FD[T_1[l_1(x),i], T_2[l_2(y),j]] \leftarrow cost$
23:                 $D[i,j] \leftarrow cost$
24:                 $FDPATH[T_1[l_1(x),i], T_2[l_2(y),j]] \leftarrow path$
25:                 $DPATH[i,j] \leftarrow path$
26:                 **else**
27:                     $cost, path \leftarrow best(\{FD[T_1[l_1(x),i\text{-}1], T_2[l_2(y),j]] + \gamma(T_1[i] \longrightarrow \wedge),$
28:                               $FDPATH[T_1[l_1(x),i\text{-}1], T_2[l_2(y),j]] +$ "d"$\},$
29:                             $\{FD[T_1[l_1(x),i], T_2[l_2(y),j\text{-}1]] + \gamma(\wedge \longrightarrow T_2[j]),$
30:                             $FDPATH[T_1[l_1(x),i], T_2[l_2(y),j\text{-}1]] +$ "i"$\},$
31:                             $\{FD[T_1[l_1(x),i\text{-}1], T_2[l_2(y),j\text{-}1]] + D[i,j]),$
32:                             $FDPATH[T_1[l_1(x),i\text{-}1], T_2[l_2(y),j\text{-}1]] + DPATH[i][j]\})$
33:                 $FD[T_1[l_1(x),i], T_2[l_1(y),j]] \leftarrow cost$
34:                 $FDPATH[T_1[l_1(x),i], T_2[l_1(y),j]] \leftarrow path$
35:             **end if**
36:             **end for**
37:         **end for**
38:     **end for**
39: **end for**
40: **return** $D[n,m], DPATH[n,m]$

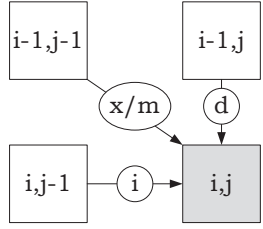

Figure 3: The edit operation direction used in our algorithm. Each arc that implies an edit operation is labeled: "i" for an insertion, "d" for deletion, "x" for exchanging and "m" for no operation (matching).

the optimal sequence of edit operations that transform $T_1$ into $T_2$ are highlighted in bold, with the final optimal path shown in the last cell (at final row and column).

| $T_2$ | | g | c | y | z | x | d | a |
|---|---|---|---|---|---|---|---|---|
| $T_1$ | - | i | ii | iii | iiii | iiiii | iiiiii | iiiiiii |
| e | **d** | x | xi | iix | iiix | iiixi | iiixii | iiixiii |
| f | **dd** | xd | xid | xix | iixx | xiiix | xiiiix | xiiiixi |
| b | **ddd** | xdd | xdx | xdxi | iixxd | iixxx | iixxxi | iixxxii |
| g | dddd | **dddm** | dddmi | xdxx | xdxxi | xdxxii | xdxixii | iixxxiid |
| c | ddddd | dddmd | **dddmm** | **dddmmi** | **dddmmii** | **dddmmiii** | xdxixix | xdxixixi |
| d | dddddd | dddmdd | dddmmd | dddmmx | dddmmxi | dddmmxii | **dddmmiiim** | dddmmiiimi |
| a | ddddddd | dddmddd | dddmmdd | dddmmxd | dddmmxid | dddmmxiid | dddmmiiimd | **dddmmiiimm** |

FDPATH

Figure 4: Computing the optimal path for the trees in Figure 2.

The mapping between two trees can be found from the final sequence of edit operations by mapping the nodes corresponding to match operation "m" only.

The final distance is 6 which represents the final values (at final row and column) in $D$.[2] The last value in $DPATH$ represents the final sequence of edit operations, namely **dddm-miiimm**. According to this path, we can define an *alignment* between two postorder trees. The alignment between two trees $T_1$ and $T_2$ is obtained by inserting a *gap symbol* (i.e. "_") into either $T_1$ or $T_2$, according to the type of edit operation, so that the resulting strings $S^1$ and $S^2$ are the same length as the sequence of edit operations. The gap symbol is inserted into $S^2$ when the edit operation is delete ("d"), whereas it is inserted in $S^1$ when the edit operation is insert ("i"). Otherwise, the nodes of $T_1$ and $T_2$ are inserted into $S^1$ and $S^2$ respectively. The following is an optimal alignment between $T_1$ and $T_2$:

$$S^1: \quad e \quad f \quad b \quad g \quad c \quad \_ \quad \_ \quad \_ \quad d \quad a$$
$$\quad\quad \mathbf{d} \quad \mathbf{d} \quad \mathbf{d} \quad \mathbf{m} \quad \mathbf{m} \quad \mathbf{i} \quad \mathbf{i} \quad \mathbf{i} \quad \mathbf{m} \quad \mathbf{m}$$
$$S^2: \quad \_ \quad \_ \quad \_ \quad g \quad c \quad y \quad z \quad x \quad d \quad a$$

---

2. For simplicity here, we assume that the each single operation will cost 1 except that matching will cost 0, as described by Zhang and Shasha (1989).

This means:

| | |
|---|---|
| **d:** | Delete $(e)$ from $T_1$ |
| **d:** | Delete $(f)$ from $T_1$ |
| **d:** | Delete $(b)$ from $T_1$ |
| **m:** | Leave $(g)$ without change |
| **m:** | Leave $(c)$ without change |
| **i:** | Insert $(y)$ into $T_1$ |
| **i:** | Insert $(z)$ into $T_1$ |
| **i:** | Insert $(x)$ into $T_1$ |
| **m:** | Leave $(d)$ without change |
| **m:** | Leave $(a)$ without change |

The final mapping between $T_1$ and $T_2$ is shown in Figure 5. For each mapping figure the insertion, deletion, matching and exchanging operations are shown with single, double, single dashed and double dashed outline respectively. The matching nodes (or subtrees) are linked with dashed arrows.

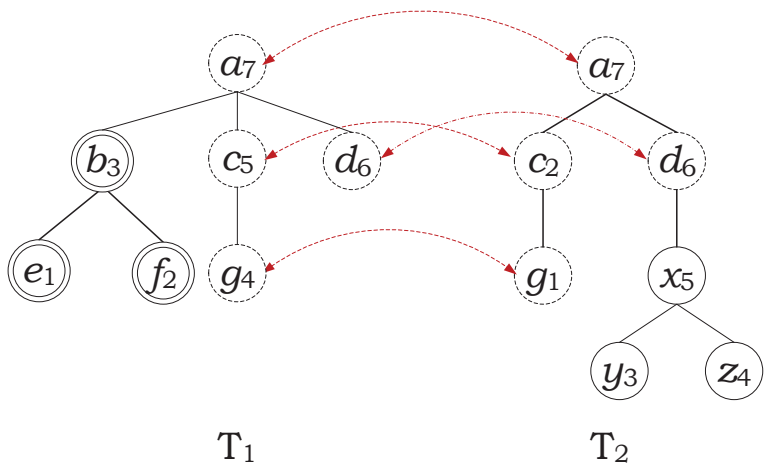

Figure 5: ZS-TED, mapping between $T_1$ and $T_2$.

## 3. Extended TED with Subtree Operations

The main weakness of the ZS-TED algorithm is that it is not able to perform transformations on subtrees (i.e. delete subtree, insert subtree and exchange subtree). The output of ZS-TED is the lowest cost sequence of operations on single nodes. We extend this to find the lowest cost sequence of operations on nodes and subtrees, TED+ST, as follows:

1. Run ZS-TED and compute the standard alignment from the results (Algorithm 1);

2. Go over the alignment and group subtree operations. Where a sequence of identical operations applies to a set of nodes comprising a subtree, they are replaced by a single operation, whose cost is determined by some appropriate function of the costs of the individual nodes (Algorithm 2). A variety of functions could be applied here, depending on the application. When using the algorithm for textual entailment we use the costs in Figure 8, which are derived from those used by Punyakanok, Roth, and Yih (2004), but for illustration in the current section we will simply take the cost of a subtree operation to be half the sum of the costs of the individual operations that make it up.

It should be noted here that while we apply this technique to modify Zhang-Shasha's $O(n^4)$ algorithm, it could also be applied to any other algorithm for finding tree edit distance, e.g. Klein's $O(n^3 log_n)$ algorithm (Klein, 1998), Demaine et al. $O(n^3)$ algorithm (Demaine, Mozes, Rossman, & Weimann, 2009) or Pawlik and Augsten $O(n^3)$ algorithm (Pawlik & Augsten, 2011), since the extension operates on the output of the original algorithm. The additional time cost of $O(n^2)$ is negligible since it is less than the time cost for any available tree edit distance algorithm.

## 3.1 Find a Sequence of Subtree Edit Operations

Extending ZS-TED to cover subtree operations will give us more flexibility when comparing trees (especially linguistic trees). The key to this algorithm is that we have to find maximal sequences of identical edit operations which correspond to subtrees. A sequence of nodes in postorder corresponds to a subtree if the following conditions are satisfied: (i) the first node is a leaf; and (ii) the leftmost sibling of the last node in the sequence (i.e. the root of a subtree) is the same as the first node in the sequence. These two conditions can be checked in constant time, since the leftmost sibling of a node can determined for each node in advance. We can hence find maximal sequences corresponding to subtrees by scanning forwards through the sequence of node operations to find sequences of identical operations, and then scanning backwards through such a sequence until we find the point at which it covers a subtree. This involves potentially $O(n^2)$ steps–$n$ forward steps to find sequences of identical operations, and then possibly $n$-1 backward steps each time to find sub-sequences corresponding to subtrees. As an example, the sequence of nodes $e,f,b$ in tree $T_1$ in Figure 2 is a subtree because $e$ is a leaf and the leftmost of the last node $b$ is 1, which represents the first node $e$. On the other hand, the sequence of nodes $g,c,d$ in the same tree is not a subtree because $g$ is a leaf, but the leftmost of the last node $d$ is 6, which represents itself, not the first node $g$.

Algorithm 2 contains the pseudocode to find the optimal sequence of single and subtree edit operations for transforming $T_1$ into $T_2$. $E_{p=1..L} \in \{$"d", "i", "x", "m"$\}$ in this algorithm is an optimal sequence of node edits for transforming $T_1$ into $T_2$, obtained by applying the technique in Section 2, and $S^1$ and $S^2$ are the alignments for $T_1$ and $T_2$ obtained after applying this sequence of node edits.

As shown in Algorithm 2, to find the optimal single and subtree edit operations sequence that transforms $T_1$ into $T_2$, each maximal sequence of identical operations is checked to see whether it contains subtree(s) or not. Checking whether such a sequence corresponds to a subtree depends on the type of edit operation, according to the following rules: (i) if the operation is "d," the sequence is checked on the first tree; (ii) if the operation is "i," the sequence is checked on the second tree; and (iii) otherwise, the sequence is checked on both

---

**Algorithm 2** pseudocode to find subtree edit operations

| | |
|---|---|
| E | the sequence of edit operations that transform tree $T_1$ into tree $T_2$, $E_{p=1..L} \in \{\text{"d","i","x","m"}\}$ |
| L | the length of the sequence of edit operations E |
| $S^1, S^2$ | the optimal alignment for $T_1$ and $T_2$ respectively, when the length of $S^1 = S^2 = L$ |

1: **repeat**
2:     $ERoot \leftarrow E_L$
3:     $F \leftarrow L$
4:     **repeat**
5:         **while** $(F \geq 2$ **and** $E_{F-1} == ERoot)$ **do**
6:             $F \leftarrow F - 1$
7:         **end while**
8:         **if** $(F == L)$ **then**
9:             $L \leftarrow L - 1$
10:            $ERoot \leftarrow E_L$
11:            $F \leftarrow L$
12:         **end if**
13:     **until** $(F < L$ **and** $F \geq 2$ **and** $E_{F-1} \neq ERoot)$ **or** $(L = 0)$
14:     $F0 \leftarrow F$
15:     **while** $(F < L)$ **do**
16:         **while** $(F < L)$ **do**
17:             $IsSubtree \leftarrow true$
18:             **while** $(F < L$ **and** $IsSubtree)$ **do**
19:                 **if** $(ERoot =$"d" **and** $S_F^1..S_L^1$ *is subtree*$)$ **or**
20:                 $(ERoot =$"i" **and** $S_F^2..S_L^2$ *is subtree*$)$ **or**
21:                 $((ERoot$ *in* $\{$"x","m"$\})$ **and** $(S_F^1..S_L^1$ **and** $S_F^2..S_L^2$ *are subtrees*$))$ **then**
22:                     *Replace* $E_F..E_{L-1}$ *with* "+"
23:                     $L \leftarrow F - 1$
24:                     $F \leftarrow F0$
25:                 **else**
26:                     $IsSubtree \leftarrow false$
27:                 **end if**
28:             **end while**
29:             $F \leftarrow F + 1$
30:         **end while**
31:         $L \leftarrow L - 1$
32:         $F \leftarrow F0$
33:     **end while**
34:     $L \leftarrow F0 - 1$
35: **until** $(L \leq 0)$
36: **return**   $E$

---

trees. After that, if the sequence of operations corresponds to a subtree, then all the symbols of the sequence are replaced by "+" except the last one (which represents the root of the subtree). Otherwise, checking starts from a sub-sequence of the original, as explained below. For instance, let us consider $E_h, ..., E_t$, where $1 \leq h < L$, $1 < t \leq L$, $h < t$, is a sequence of the *same* edit operation, i.e. $E_{k=h..t} \in \{\text{"d", "i", "x", "m"}\}$. Let us consider $h0 = h$, we firstly check nodes $S_h^1, ..., S_t^1$ and $S_h^2, ..., S_t^2$ to see whether or not they are the heads of subtrees. If $E_k$ is *"d,"* the nodes $S_h^1, ..., S_t^1$ are checked, if it is *"i"* the nodes $S_h^2, ..., S_t^2$ are checked, and otherwise, the nodes $S_h^1, ..., S_t^1$ and $S_h^2, ..., S_t^2$ are checked. All edit operations $E_h, ..., E_{t-1}$ are replaced by "+" when this sequence corresponds to a subtree. Then, we start checking from the beginning of another sequence from the left of the subtree $E_h, ..., E_t$, i.e. $t = h - 1$.

Otherwise, the checking is applied with the sequence starting from the next position, i.e. $h = h + 1$. The checking is continued until $h = t$. After that, when the $(t - h)$ sequences that start with different positions and end with $t$ position do not contain a subtree, the checking starts from the beginning with the new sequence, i.e. $h = h0$ and $t = t - 1$. The process is repeated until $h = t$.

To explain how the subtree operations are applied, let us consider the two trees $T_1$ and $T_2$ in Figure 2.

According to TED+ST, the cost is 3 and the sequence of operation is as follows: there is a sequence of "**d**," "**m**" and "**i**" in the result. These sequences consist of three subtrees (i.e. the three deleted nodes, the first two matched nodes and the three inserted nodes): **<u>ddd</u> <u>mm</u> <u>iii</u> mm**. So, the final result is: <u>++**d**</u> <u>+**m**</u> <u>++**i**</u> **mm**. This means:

    **++d:**    Delete subtree $(e,f,b)$ from $T_1$
    **+m:**    Leave subtree $(g,c)$ without change
    **++i:**    Insert subtree $(y,z,x)$ into $T_1$
    **m:**    Leave $(d)$ without change
    **m:**    Leave $(a)$ without change

The final mapping between $T_1$ and $T_2$ obtained using TED+ST is shown in Figure 6.

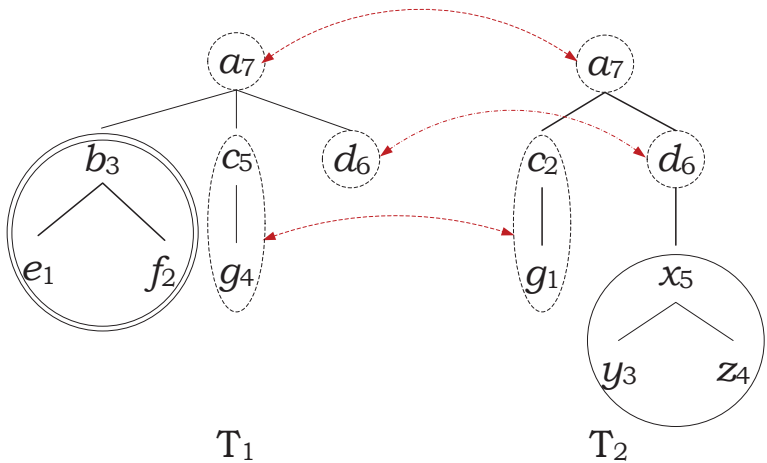

Figure 6: TED+ST, mapping between $T_1$ and $T_2$.

## 4. Matching Dependency Trees

As mentioned above, our main goal is to design a *textual entailment* (TE) system for Arabic to check whether one text snippet (i.e. premise $p$) entails another text (i.e. hypothesis $h$). To match $p$ and $h$ dependency tree pairs effectively, we use TED+ST. This enables us to find the minimum edit operations to transform one tree to another. This allows us to be sensitive to the fact that the links in a dependency tree carry linguistic information about relations between complex units, and hence to ensure that we are paying attention to these relations when we compare two trees. For instance, this enables us to pay attention to the fact that operations involving modifiers, in particular, should be applied to the subtree as a whole rather than to its individual elements. Thus, we transform tree $D_1$ to tree $D_2$ in Figure 7 by deleting "in the park" in a single operation, removing the modifier as a whole, rather than three operations removing "in," "the" and "park" one by one, using the costs in Figure 8 as an initial test for edit operations in our experiments. These costs are an updated version of the costs used by Punyakanok et al. (2004).[3] These authors found that using tree edit distance gives better results than bag-of-word scoring methods, when they applied them for question answering.[4]

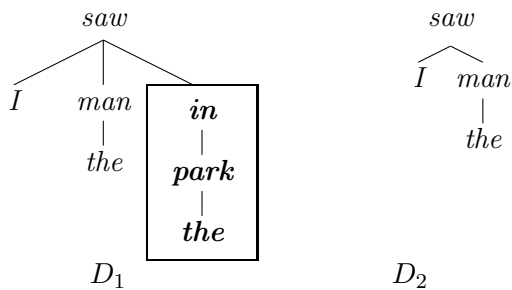

Figure 7: Two dependency trees, $D_1$ and $D_2$.

By using the costs in Figure 8, the cost of transferring $D_1$ into $D_2$ according to ZS-TED is 19 (i.e. one stop word "the" (5) and two words (14)), whereas according to TED+ST operations it is 0. Therefore, it is easy to decide that $D_1$ entails $D_2$, whereas the reverse is not true. We also exploited the subset/superset relations encoded by Arabic WordNet (AWN) (Black, Elkateb, Rodriguez, Alkhalifa, Vossen, Pease, & Fellbaum, 2006) when comparing items in a tree. Roughly speaking, if comparing one tree to another requires us to swap two lexical items, we will be happier doing so if the item in the source tree is a synonym or hyponym of the one in the target tree–since "wombat" is a hyponym of "animal," swapping "wombat" in a premise such as "I saw a wombat at the zoo" for "animal" in "I saw an animal at the zoo" is a truth-preserving exchange.

Approaches that make use of lexical relations of this kind have to cope with the fact that words often have multiple meanings. We follow Hobbs (2005) in assuming that if $W_1$ has a sense which is a hyponym of some sense of $W_2$ then a sentence involving $W_1$ will entail a similar sentence involving $W_2$ as shown in (1).

(1) *p.*    I saw a *peach* at yesterday's party.

   *h.*    I saw a *very attractive woman* at yesterday's party.

---

3. The stop words here are a list that contains some of the most common Arabic words (e.g. the particle إن *Ǎn* "indeed"). For instance, إن المدير مشغول *Ǎn Almdyr mšγwl* "The director is **indeed** busy" entails المدير مشغول *Almdyr mšγwl* "The director is busy."

4. The transcription of Arabic examples in this document follows Habash-Soudi-Buckwalter (HSB) transliteration scheme (Habash, Soudi, & Buckwalter, 2007) for transcribing Arabic symbols.

| Cost | Single node | Subtree (more than one node) |
|---|---|---|
| **Delete:** | if X is a stop word then cost is 5, else cost is 7 | 0 |
| **Insert:** | if a Y is a stop word then cost is 5, else cost is 100 | double the sum of the costs of its parts |
| **Exchange:** | if X is subsumed by Y cost is 0, elseif X is a stop word cost is 5, elseif Y is subsumed by (or is an antonym of) a X then cost is 100 else cost is 50 | if S1 is identical to S2 then cost is 0 else cost half the sum of the costs of its parts |

Figure 8: Edit operation costs.

In (1$p$), for instance, the word "peach" is ambiguous,[5] "a shade of pink tinged with yellow" (hypernym: Pink) or "Downy juicy fruit with sweet yellowish or whitish flesh" (hypernym: Drupe, edible fruit, stone fruit) or "a very attractive or seductive looking woman" (hypernym: Adult female, women) or "cultivated in temperate regions" (hypernym: Fruit tree). In the context of (1$h$), however, any human reader would assume that the second interpretation of "peach" was intended, despite the fact that it is in general a fairly unusual usage.

This reflects the widely accepted view that contextual information is the key to lexical disambiguation. Within the RTE task, the premise provides the context for disambiguation of the hypothesis, and the hypothesis provides the context for disambiguation of the premise. Almost any human reader would, for instance, accept that (2$p$) entails (2$h$), despite the potential ambiguity of the word "bank."

(2)   $p.$   My money is all tied up at the bank.

     $h.$   I cannot easily spend my money.

## 5. Dataset Preparation

In order to train and test our TE system for Arabic, we need an appropriate dataset. To our knowledge, no such datasets are available for Arabic, so we have had to develop one. We have followed one of the procedures used for collecting the premise-hypothesis pairs in the RTE tasks, with a slight alteration. The premises in RTE were collected from a variety of sources, e.g. newswire text. They contain one or two sentences and tend to be fairly long (e.g. averaging 25 words in RTE1, 28 words in RTE2, 30 words in RTE3 and 39 words in RTE4). In contrast, the hypotheses are quite short single sentences (averaging 11 words in RTE1, 8 words in RTE2 and 7 words in RTE3 and RTE4), which were manually constructed for each premise. The first three RTE Challenges were presented as a binary classification task 'yes' or 'no' with balanced numbers of 'yes' and 'no' problems. Beginning with RTE4, there were three-way classifications ('yes,' 'no,' or 'contradict,' to distinguish cases in which $h$ contradicts $p$ from those in which $h$ is compatible with, but not entailed by $p$). In our dataset, we do not want to produce a set of *p-h* pairs by hand–partly because doing so is a lengthy and tedious process, but more importantly because hand-coded datasets are liable to embody biases introduced by the developer. If the dataset is used for training the system, then the rules that are extracted will be little more than an unfolding of information explicitly supplied by the developers. If it is used for testing then it will only test the examples that the developers have chosen, which are likely to be biased, albeit unwittingly, towards the way they think about the problem.

---

5. See Sage's dictionary online: `http://www.sequencepublishing.com/thesageonline.php`. WordNet also provides all these senses (and more) for "peach."

Our set of Arabic *p-h* pairs for the TE task was created by a semi-automatic technique through two stages. The first stage (Section 5.1) is responsible for automatically collecting *p-h* pairs from news websites, while the second stage (Section 5.2) uses an online annotation system that allows annotators to annotate our collected pairs manually. Both stages are explained in detail below.

## 5.1 Collecting *p-h* Pairs

We collected candidate *p-h* pairs automatically by the so-called *headline-lead paragraph technique* (Burger & Ferro, 2005) from the web (e.g. from newspaper corpora, pairing the first paragraph of article, as *p*, with its headline, as *h*). This is based on the observation that a news article's headline is very often a partial paraphrase of the first paragraph of this article, conveying thus a comparable meaning. We use an updated version of the headline-lead paragraph strategy to improve the quality of the *p-h* pair.

The key idea here is that we pose queries to a search engine and automatically filter the responses for text snippets that might entail the query. These pairs are then manually annotated for entailment/non-entailment, but the texts themselves are automatically collected from freely occurring natural texts. This eliminates the possibility (indeed likelihood) of unconscious bias that is introduced if the hypotheses are manually generated.

We built a corpus of *p-h* pairs by using headlines from the websites of Arabic newspapers and TV channels as queries to be input to Google via the standard Google API, and then selecting the first paragraph, which usually represents the most related text snippet(s) in the article with the headline (Burger & Ferro, 2005), of each of the first 10 returned pages. This technique produces a large number of potential pairs without any bias in either the premises or the hypotheses. To improve the quality of the pairs that resulted from the query, we use two conditions to filter the results: (i) the length of a headline must be at least five words, to avoid very small headlines; and (ii) fewer than 80% of the words in the headline should appear in the premise, to avoid having very similar sentences.

The problem here is that if *p* and *h* are very similar then there would be very little to learn from them if they were used in the training phase of a TE system; and they would be almost worthless as a test pair–virtually any TE system will get such a pair right, so they will not serve as a discriminatory test pair. We therefore eliminate excessively similar *p-h* pairs from both training and testing, which we assess in terms of the number of shared uncommon words.

In order to overcome this problem, we matched headlines from one source with stories from another. Major stories are typically covered by a range of outlets, usually with variations in emphasis or wording. Stories from different sources can be linked by looking for common words in the headlines–it is unlikely that there will be two stories about, for in-stance, neanderthals in the news at the same time, so very straightforward matching based on low frequency words and proper names is likely to find articles about the same topic. The terminology and structure of the first text snippets of these articles, however, are likely to be quite different. Thus using a headline from one source and the first text snippet from an article about the same story but from another source is likely to produce *p-h* pairs which are not unduly similar. We can therefore link a headline from one newspaper with related sentences from another.

## 5.2 Annotating *p-h* Pairs

The pairs that are collected in the first stage still have to be marked-up by human annotators, but at least the process of collecting them is as nearly bias-free as possible. These pairs cover a number of subjects such as politics, business, sport and general news. The annotation is performed by eight expert and non-expert human annotators to identify the different pairs as positive entailment examples 'yes,' where *p* is judged to entails *h*, and as negative examples 'no,' where entailment does not hold. Those annotators follow nearly the same annotation guidelines as those used for building the RTE task dataset (Dagan, Glickman, & Magnini, 2006).

Each pair was annotated by three annotators. The inter-annotator agreement (where all annotators agree) is around 74% compared with 89% where each annotator agrees with at least one co-annotator. This suggests that the annotators found this is a difficult task. The fact that there was only 74% agreement when the annotations produced by three independent annotators are taken into account sets an upper bound on what it is reasonable to expect of an automatic system for carrying out this task. If human annotators can only agree in about three quarters of the cases, then it is unlikely that a computer-based system can achieve much more than 75% agreement with any given pair of annotators.[6]

## 6. Experiments

To check the effectiveness of TED+ST, we used it to check the entailment between *p-h* Arabic pairs of text snippets and compared its results with two string-based approaches (bag-of-words and Levenshtein distance) and ZS-TED on the same set of pairs. Checking whether one Arabic text snippet entails another, however, is particularly challenging because Arabic is more ambiguous than most languages, such as English. For instance, Arabic is written without diacritics (short vowels), often leading to multiple ambiguities. This makes morphological analysis very difficult (i.e. a single written form may easily correspond to as many as ten different lexemes, see Alabbas & Ramsay, 2011a, 2011b, 2012a, 2012c). The preliminary testing dataset contains 600 pairs, binary annotated as 'yes' and 'no' (a 50-50 split) using the technique explained in Section 5. The distribution of these pairs over *p* length is summarised in Table 1, when the *h* average length is around 10 words and the average of common words between *p* and *h* is around 4 words. The average length of sentence in this dataset is 25 words per sentence, with some sentences containing 40+ words.

---

6. The dataset, including the dependency-tree analysis in CoNLL format, is available in the online appendices to this article or from `http://www.cs.man.ac.uk/~ramsay/ArabicTE/`

| *p's* length | #pairs | yes | no |
|:---|:---:|:---:|:---:|
| <20 | 175 | 83 | 92 |
| 20-29 | 329 | 171 | 158 |
| 30-39 | 87 | 43 | 44 |
| >39 | 9 | 3 | 6 |
| **Total** | **600** | **300** | **300** |

Table 1: Distribution of sentence lengths in the testset.

In order to check the entailment between *p-h* pairs, we follow three steps. First, each sentence is preprocessed by a tagger and a parser in order to convert both elements of the *p-h* pair to dependency trees. A dependency tree is a tree where words are vertices and syntactic relations are dependency relations. Each vertex therefore has a single parent, except the root of the tree. A dependency relation holds between a *dependent*, i.e. a syntactically subordinate vertex, and a *head*, i.e. another vertex on which it is dependent. Thus the dependency structure is represented as a head-dependent relation between vertices that are classified by dependency types such as `SBJ` "subject," `OBJ` "object," `ATT` "attribute," etc.

We have carried out a number of experiments with state-of-the-art taggers such as AMIRA (Diab, 2009), MADA (Habash, Rambow, & Roth, 2009) and an in-house maximum-likelihood (MXL) tagger (Ramsay & Sabtan, 2009) and parsers such as MALTParser (Nivre, Hall, Nilsson, Chanev, Eryigit, Kübler, Marinov, & Marsi, 2007) and MSTParser (McDonald, Lerman, & Pereira, 2006).[7] These experiments show in particular that merging MADA (97% accuracy) with MSTParser gives better results (around 81% for labelled accuracy) than the other tagger:parser combinations (Alabbas & Ramsay, 2012b). We therefore use MADA+MSTParser in the current experiments.

After converting *p-h* pairs to dependency trees, we matched these dependency trees using the ZS-TED and TED+ST algorithms, with two string-based algorithms (bag-of-words and Levenshtein distance) to provide a baseline. The tree edit distance algorithms used the edit operation costs defined in Figure 8 to find the cost of matching between *p-h* pairs. The bag-of-words here measures the similarity between *p* and *h* as a number of common words between them (either in surface forms or lemma forms), divided by the length of *h*. For all four algorithms we use AWN as a lexical resource in order to take account of synonymy and hyponymy relations when calculating the cost of an edit.

We carried out two kinds of experiments using these algorithms: the first was a simple yes/no experiment, using a single threshold to decide whether the premise was similar enough to the hypothesis for it to be safe to say that it entailed it, and the second with two thresholds so that we could say yes/don't know/no. The results of these experiments are given below.

---

7. These parsers are data-driven dependency parsers. For Arabic they are usually trained on an Arabic dependency treebank, such as Prague Arabic Dependency Treebank (PADT) (Smrž, Bielicky, Kouřilová, Kráčmar, Hajič, & Zemánek, 2008), or on some version of the Penn Arabic Treebank (PATB) (Maamouri & Bies, 2004) that has been converted to dependency trees: scoring of such parsers is a matter of counting dependency links.

### 6.1 Binary Decision ('yes' and 'no')

$p$ entails $h$ when the cost of matching is less (more in case of bag-of-words) than a threshold. The results of these experiments, in terms of precision (P), recall (R) and F-score (F) for 'yes' class and overall accuracy, are shown in Table 2. This table shows the substantial improvement obtained by using TED+ST over the bag-of-words (F-score for TED+ST is around 1.16 times the F-score for bag-of-words, and accuracy is about 1.09 times better) and ZS-TED (around 1.06 times better in F-score and 1.04 times better in total accuracy).

| Method | $P_{yes}$ | $R_{yes}$ | $F_{yes}$ | Accuracy |
|---|---|---|---|---|
| Bag-of-words | 63.5% | 43.7% | 0.518 | 59.3% |
| Levenshtein distance | 64.7% | 44.1% | 0.525 | 60.2% |
| ZS-TED | 65.9% | 51.2% | 0.576 | 62.5% |
| TED+ST | 69.7% | 54.5% | 0.612 | 65.5% |

Table 2: Performance of TED+ST compared with the string-based algorithms and ZS-TED, binary decision.

Although we are primarily interested in Arabic, we have carried out parallel sets of experiments on the English RTE2 testset, using the Princeton English WordNet (PWN) as a resource for deciding whether a word in the premise may be exchanged for one in the hypothesis. Because the tree edit distance algorithms work with dependency tree analyses of the input texts, we have used a set that have been analysed using Minipar (Lin, 1998), downloaded from `http://u.cs.biu.ac.il/~nlp/RTE2 Datasets/RTE2 Preprocessed Datasets.html`. The RTE2 testset contains around 800 $p$-$h$ pairs, but a number of the Minipar analyses have multiple heads and hence do not correspond to well-formed trees, and there are also a number of cases where the segmentation algorithm that was used produces multi-word expressions. After eliminating problematic pairs of this kind we are left with 730 pairs, split evenly between positive and negative examples. Since we are mainly concerned here with the difference between ZS-TED and TED+ST, we have omitted the Levenshtein distance and have simply kept the basic bag-of-words algorithm as a baseline. Previous authors have shown that tree edit distance consistently outperforms string-based approaches on this dataset, and there is no need to replicate that result here.

| Method | $P_{yes}$ | $R_{yes}$ | $F_{yes}$ | Accuracy |
|---|---|---|---|---|
| Bag-of-words | 53.2% | 50.1% | 0.516 | 52.1% |
| ZS-TED | 52.9% | 62.5% | 0.573 | 53.5% |
| TED+ST | 53.2% | 66.8% | 0.59 | 55.8% |

Table 3: Performance of TED+ST compared with the simple bag-of-words and ZS-TED, binary decision, RTE2 dataset.

The pattern in Table 3 is similar to that in Table 2. ZS-TED is better than bag-of-words, TED+ST is a further improvement over ZS-TED. Most experiments on textual entailment tasks only report accuracy: in certain situations it may be more important to have decisions that are trustworthy (high precision, as in Table 2) or to be sure that you have captured as many positive examples as possible (high recall[8], as in Table 3), or to have a good balance between these (high F-score). It is easy to change the balance between precision and recall, simply by changing the threshold that is used for determining whether it is safe to say that $p$ entails $h$–we could have chosen thresholds for Table 3 that increased the precision and decreased the recall, so that the results more closely matched Table 2. The key point here is that in both sets of experiments, the F-scores improve as we move from string-based measures to ZS-TED and then again when we use TED+ST; and that they are remarkably similar for the two datasets, despite the fact that they were collected by different means, are in different languages, and are parsed using different parsers.

### 6.2 Making a Three-way Decision ('yes,' 'no' and 'unknown')

For this task we use two thresholds, one to trigger a positive answer if the cost of matching is lower than the lower threshold (exceeds the higher one for the bag-of-words algorithm) and the other to trigger a negative answer if the cost of matching exceeds the higher one (*mutatis mutandis* for bag-of-words). Otherwise, the result will be 'unknown.' The reason for making a three-way decision is to drive systems to make more precise distinctions. Note that we are not distinguishing here between {$h$ entails $p$, $h$ and $p$ are compatible, $h$ contradicts $p$}, but between {$h$ entails $p$, I don't know whether $h$ entails $p$, $h$ does not entail $p$}. This is a more subtle distinction, reflecting the system's confidence in its judgement, but it can be extremely useful when deciding how to act on its decision.

The results of this experiment, in terms of precision (P), recall (R) and F-score (F), are shown in Table 4. Again, it shows the large improvement of using TED+ST over the bag-of-words (F-score is around 1.10 times better) and ZS-TED (F-score around 1.06 times better).

| Method | P | R | F |
|---|---|---|---|
| Bag-of-words | 58.9% | 56.7% | 0.578 |
| Levenshtein distance | 61.4% | 58.0% | 0.597 |
| ZS-TED | 65.1% | 56.0% | 0.602 |
| TED+ST | 67.4% | 60.2% | 0.636 |

Table 4: Performance of TED+ST compared with string-based algorithms and ZS-TED, three-way decision.

The scores for the three-way decision on the RTE2 dataset are lower than for our Arabic dataset, but again TED+ST outperforms ZS-TED on all three measures.

### 7. Conclusion

We have presented here an extended version, TED+ST, of tree edit distance that solved one of the main drawbacks of standard tree edit distance, which is that it only supports edit operations (i.e. delete, insert and exchange) on single nodes. TED+ST deals with subtree transformation operations as well as operations on single nodes: this leads to useful improvements over the performance of the standard algorithm for determining entailment. The key here is that subtrees tend to correspond to single information units. By treating operations on subtrees as less costly than the corresponding set of individual node operations, TED+ST concentrates on entire information units, which are a more appropriate granularity than individual words for considering entailment relations.

---

8. This might be useful, for instance, with TED+ST being used as a low cost filter in a question-answering system, where the results of a query to a search engine might be filtered by TED+ST before being passed to a system employing full semantic analysis and deep reasoning, which are high precision but are also very time-consuming.

| Method | P | R | F |
|---|---|---|---|
| Bag-of-words | 50.8% | 48.3% | 0.495 |
| ZS-TED | 52.3% | 50.2% | 0.512 |
| TED+ST | 54.3% | 52.7% | 0.535 |

Table 5: Performance of TED+ST compared with the simple bag-of-word and ZS-TED, three-way decision, RTE2 dataset.

The current findings, while preliminary, are quite encouraging. The fact that the results on our original testset, particularly the improvement in F-score, were replicated for a testset where we had no control over the parser that was used to produce dependency trees from the $p$-$h$ pairs provides some evidence for the robustness of the approach. We anticipate that in both cases having a more accurate parser (our parser for Arabic attains around 81% accuracy on the PATB, Minipar is reported to attain about 80% on the Suzanne corpus) would improve the performance of both ZS-TED and TED+ST.

We are currently experimenting with different scoring algorithms for ZS-TED and TED+ST. The performance of any variant of tree edit distance depends critically on the costs for the various operations, and on the thresholds that are used for deciding whether $h$ entails $p$, and we are therefore investigating the use of various optimisation algorithms for choosing these weights and thresholds. We also intend to use other Arabic lexical resources, such as OpenOffice Arabic dictionary and MS Word Arabic dictionary, to provide us with more information about relations between words, because the information in AWN, while very useful, is sparse in comparison to PWN (Habash, 2010).

## Acknowledgments

We would like to thank the reviewers for their valuable comments, in particular the reviewer who suggested evaluating the approach on an English dataset as well as our Arabic one. The extra work has provided support for our belief in the robustness of the approach to a degree that we did not anticipate.

We would like to extend our thanks to our annotators for the time and effort they have put into annotating our experimental dataset. Maytham Alabbas owes his deepest gratitude to Iraqi Ministry of Higher Education and Scientific Research for financial support in his PhD study. Allan Ramsay's contribution to this work was partially supported by the Qatar National Research Fund (grant NPRP 09 - 046 - 6 - 001).

# References

Alabbas, M. (2011). ArbTE: Arabic textual entailment. In *Proceedings of the Second Student Research Workshop associated with RANLP 2011*, pp. 48–53, Hissar, Bulgaria. RANLP 2011 Organising Committee.

Alabbas, M., & Ramsay, A. (2011a). Evaluation of combining data-driven dependency parsers for Arabic. In *Proceeding of 5th Language & Technology Conference: Human Language Technologies (LTC'11)*, pp. 546–550, Poznań, Poland.

Alabbas, M., & Ramsay, A. (2011b). Evaluation of dependency parsers for long Arabic sentences. In *Proceeding of International Conference on Semantic Technology and Information Retrieval (STAIR'11)*, pp. 243–248, Putrajaya, Malaysia. IEEE.

Alabbas, M., & Ramsay, A. (2012a). Arabic treebank: from phrase-structure trees to dependency trees. In *META-RESEARCH Workshop on Advanced Treebanking at the 8th International Conference on Language Resources and Evaluation (LREC)*, pp. 61–68, Istanbul, Turkey.

Alabbas, M., & Ramsay, A. (2012b). Combining black-box taggers and parsers for modern standard Arabic. In *Federated Conference on Computer Science and Information Systems (FedCSIS-2012)*, pp. 19 –26, Wroclaw, Poland. IEEE.

Alabbas, M., & Ramsay, A. (2012c). Improved POS-tagging for Arabic by combining diverse taggers. In *Proceedings of 8th Artificial Intelligence Applications and Innovations (AIAI)*, pp. 107–116, Halkidiki, Greece. Springer.

Bille, P. (2005). A survey on tree edit distance and related problems. *Theoretical Computer Science*, *337*(1-3), 217–239.

Black, W., Elkateb, S., Rodriguez, H., Alkhalifa, M., Vossen, P., Pease, A., & Fellbaum, C. (2006). Introducing the Arabic WordNet project. In *Proceedings of the 3rd International WordNet Conference (GWC-06)*, pp. 295–299, Jeju Island, Korea.

Burger, J., & Ferro, L. (2005). Generating an entailment corpus from news headlines. In *Proceedings of the ACL Workshop on Empirical Modeling of Semantic Equivalence and Entailment*, pp. 49–54, Ann Arbor, Michigan, USA. Association for Computational Linguistics.

Dagan, I., & Glickman, O. (2004). Probabilistic textual entailment: generic applied modeling of language variability. In *PASCAL Workshop on Learning Methods for Text Understanding and Mining*, pp. 26–29, Grenoble, France.

Dagan, I., Glickman, O., & Magnini, B. (2006). The PASCAL recognising textual entailment challenge. In Quiñonero-Candela, J., Dagan, I., Magnini, B., & d'Alché Buc, F. (Eds.), *Machine Learning Challenges. Evaluating Predictive Uncertainty, Visual Object Classification, and Recognising Tectual Entailment*, Vol. 3944 of *Lecture Notes in Computer Science*, pp. 177–190. Springer Berlin, Heidelberg.

Demaine, E., Mozes, S., Rossman, B., & Weimann, O. (2009). An optimal decomposition algorithm for tree edit distance. *ACM Transactions on Algorithms (TALG)*, *6*(1), 2:1–2:19.

Diab, M. (2009). Second generation tools (AMIRA 2.0): fast and robust tokenization, POS tagging, and base phrase chunking. In *Proceedings of the 2nd International Conference on Arabic Language Resources and Tools*, pp. 285–288, Cairo, Eygpt. The MEDAR Consortium.

Habash, N. (2010). *Introduction to Arabic Natural Language Processing*. Synthesis Lectures on Human Language Technologies. Morgan & Claypool Publishers.

Habash, N., Rambow, O., & Roth, R. (2009). MADA+TOKAN: a toolkit for Arabic tokenization, diacritization, morphological disambiguation, POS tagging, stemming and lemmatization. In *Proceedings of the 2nd International Conference on Arabic Language Resources and Tools*, Cairo, Eygpt. The MEDAR Consortium.

Habash, N., Soudi, A., & Buckwalter, T. (2007). On Arabic transliteration. *Arabic Computational Morphology*, 15–22.

Heilman, M., & Smith, N. (2010). Tree edit models for recognizing textual entailments, paraphrases, and answers to questions. In *Human Language Technologies: The 2010 Annual Conference of the North American Chapter of the Association for Computational Linguistics*, pp. 1011–1019, Los Angeles, California, USA. Association for Computational Linguistics.

Hobbs, J. R. (2005). *The handbook of pragmatics*, chap. Abduction in Natural Language Understanding, pp. 724–740. Blackwell Publishing.

Klein, P. (1998). Computing the edit-distance between unrooted ordered trees. In *Proceedings of the 6th Annual European Symposium on Algorithms (ESA '98)*, pp. 91–102, Venice, Italy. Springer-Verlag.

Kouylekov, M. (2006). *Recognizing Textual Entailment with Tree Edit Distance: application to Question Answering and Information Extraction*. Ph.D. thesis, DIT, University of Trento, Italy.

Kouylekov, M., & Magnini, B. (2005). Recognizing textual entailment with tree edit distance algorithms. In *Proceedings of the1st Challenge Workshop Recognising Textual Entailment*, pp. 17–20, Southampton, UK.

Lin, D. (1998). Dependency-based evaluation of minipar. In *Workshop on the Evaluation of Parsing systems*, pp. 317–330. Springer.

Maamouri, M., & Bies, A. (2004). Developing an Arabic treebank: methods, guidelines, procedures, and tools. In *Proceedings of the Workshop on Computational Approaches to Arabic Script-based Languages*, pp. 2–9, Geneva, Switzerland.

MacCartney, B. (2009). *Natural Language Inference*. Ph.D. thesis, Department of Computer Science, Stanford University, USA.

McDonald, R., Lerman, K., & Pereira, F. (2006). Multilingual dependency parsing with a two-stage discriminative parser. In *10th Conference on Computational Natural Language Learning (CoNLL-X)*, New York, USA.

Mehdad, Y., & Magnini, B. (2009). Optimizing textual entailment recognition using particle swarm optimization. In *Proceedings of the 2009 Workshop on Applied Textual Inference (TextInfer '09)*, pp. 36–43, Suntec, Singapore. Association for Computational Linguistics.

Nivre, J., Hall, J., Nilsson, J., Chanev, A., Eryigit, G., Kübler, S., Marinov, S., & Marsi, E. (2007). MaltParser: a language-independent system for data-driven dependency parsing. *Natural Language Engineering*, *13*(02), 95–135.

Pawlik, M., & Augsten, N. (2011). RTED: a robust algorithm for the tree edit distance. *Proceedings of the VLDB Endowment*, *5*(4), 334–345.

Punyakanok, V., Roth, D., & Yih, W. (2004). Natural language inference via dependency tree mapping: An application to question answering. *Computational Linguistics*, *6*, 1–10.

Ramsay, A., & Sabtan, Y. (2009). Bootstrapping a lexicon-free tagger for Arabic. In *Proceedings of the 9th Conference on Language Engineering (ESOLEC'2009)*, pp. 202–215, Cairo, Egypt.

Selkow, S. (1977). The tree-to-tree editing problem. *Information Processing Letters*, *6*(6), 184–186.

Smrž, O., Bielicky, V., Kouřilová, I., Kráčmar, J., Hajič, J., & Zemánek, P. (2008). Prague Arabic dependency treebank: a word on the million words. In *Proceedings of the Workshop on Arabic and Local Languages (LREC 2008)*, pp. 16–23, Marrakech, Morocco.

Tai, K. (1979). The tree-to-tree correction problem. *Journal of the ACM (JACM)*, *26*(3), 422–433.

Zhang, K., & Shasha, D. (1989). Simple fast algorithms for the editing distance between trees and related problems. *SIAM Journal of Computing*, *18*(6), 1245–1262.